\documentclass[sigconf]{acmart}
\usepackage{multirow}
\AtBeginDocument{%
  }

\setcopyright{acmlicensed}
\copyrightyear{2025}
\acmYear{2025}
\setcopyright{cc}
\setcctype{by}
\acmConference[FAccT '25]{The 2025 ACM Conference on Fairness, Accountability, and Transparency}{June 23--26, 2025}{Athens, Greece}
\acmBooktitle{The 2025 ACM Conference on Fairness, Accountability, and Transparency (FAccT '25), June 23--26, 2025, Athens, Greece}
\acmDOI{10.1145/3715275.3732135}
\acmISBN{979-8-4007-1482-5/2025/06}

\acmSubmissionID{facct2025-p963}

\begin{document}

\title{Now you see it, Now you don't: Damage Label Agreement in Drone \& Satellite Post-Disaster Imagery}

\author{Thomas Manzini}
\email{tmanzini@tamu.edu}
\orcid{0009-0002-4937-1948}
\affiliation{%
  \institution{Texas A\&M University}
  \city{College Station}
  \state{Texas}
  \country{USA}
}

\author{Priyankari Perali}
\orcid{0009-0000-2735-1415}
\email{perali@tamu.edu}
\affiliation{%
  \institution{Texas A\&M University}
  \city{College Station}
  \state{Texas}
  \country{USA}
}

\author{Jayesh Tripathi}
\orcid{0009-0006-9286-195X}
\email{jtjayesh98@tamu.edu}
\affiliation{%
  \institution{Texas A\&M University}
  \city{College Station}
  \state{Texas}
  \country{USA}
}

\author{Robin Murphy}
\orcid{0000-0003-0774-4312}
\email{robin.r.murphy@tamu.edu}
\affiliation{%
  \institution{Texas A\&M University}
  \city{College Station}
  \state{Texas}
  \country{USA}
}


\begin{abstract}
  This paper audits damage labels derived from coincident satellite and drone aerial imagery for 15,814 buildings across Hurricanes Ian, Michael, and Harvey, finding 29.02\% label disagreement and significantly different distributions between the two sources, which presents risks and potential harms during the deployment of machine learning damage assessment systems. Currently, there is no known study of label agreement between drone and satellite imagery for building damage assessment. The only prior work that could be used to infer if such imagery-derived labels agree is limited by differing damage label schemas, misaligned building locations, and low data quantities. This work overcomes these limitations by comparing damage labels using the same damage label schemas and building locations from three hurricanes, with the 15,814 buildings representing 19.05 times more buildings considered than the most relevant prior work. The analysis finds satellite-derived labels significantly under-report damage by at least 20.43\% compared to drone-derived labels (p<1.2x10\textsuperscript{-117}), and satellite- and drone-derived labels represent significantly different distributions (p<5.1x10\textsuperscript{-175}). This indicates that computer vision and machine learning (CV/ML) models trained on at least one of these distributions will misrepresent actual conditions, as the differing satellite and drone-derived distributions cannot simultaneously represent the distribution of actual conditions in a scene. This potential misrepresentation poses ethical risks and potential societal harm if not managed. To reduce the risk of future societal harms, this paper offers four recommendations to improve reliability and transparency to decision-makers when deploying CV/ML damage assessment systems in practice.
\end{abstract}





%     \centering
%     \includegraphics[width=0.5\textwidth]{Facct25/Figures/rot90_suas_vs_satellite.png}
%     \caption{Three buildings with different damage labels in different views between drone (top-row) and satellite (bottom-row) aerial imagery. [Left Column] Label change due to roof repairs. [Middle Column] Label change due to debris removal. [Right Column] Label change due to changing flood waters.} %TODO: Perhaps reformat this to take up less space
%     \label{fig:suas-vs-satellite}
\maketitle

\section{Introduction}
\label{sec:introduction}
\begin{figure*}
    \centering
    \includegraphics[width=0.95\textwidth]{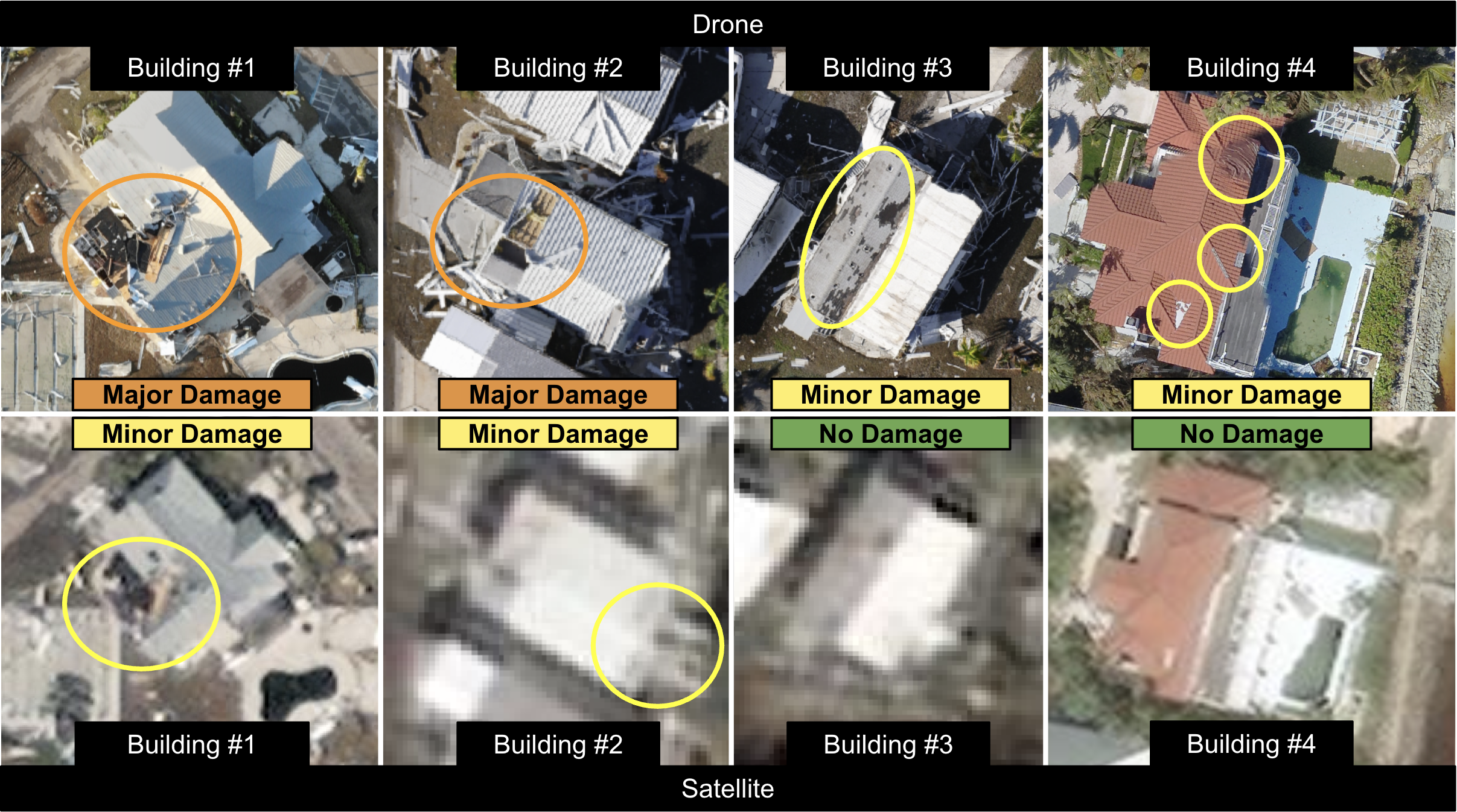}
    \caption{Four buildings with damage labels that vary between drone (top) and satellite (bottom) aerial imagery. Building damage labels are applied based on the Joint Damage Assessment (JDS) schema \cite{gupta2019creating}.} %[Building \#1] Label change due to change in resolution obscuring the structural damage. [Building \#2] Label change due to change in resolution obscuring the structural damage. [Building \#3] Label change due to change in resolution obscuring the damage to the roof. [Building \#4] Label change due to change in resolution obscuring the damage to the roof.
    \label{fig:suas-vs-satellite}
\end{figure*}

Satellite and drone imagery have become commonplace in disaster response and recovery \cite{LozanoIJDRR:23}. Aerial imagery offers substantial benefits for rapid, accurate damage assessment, as the images cover large areas often inaccessible on the ground during the immediate response and recovery phase. The disadvantage is that aerial assets produce unmanageably high volumes of imagery for manual damage assessment, arguing for the application of machine learning \cite{AnkerAOR:19,garcia2021computer,manzini2023harnessing}.

Machine learning is being explored for automating building damage assessment from aerial imagery; however, these efforts are based on an implicit assumption that labels derived only from aerial imagery can represent the actual distribution of damage present in a disaster scene. At a disaster, the ground truth for building damage is the official damage assessment label for individual buildings as assigned by an inspector on the ground physically examining a building. These official labels are often unavailable due to privacy and sharing restrictions between agencies, so there is a practical advantage to learning without training or conditioning on ground-level damage assessment labels. Not surprisingly, all known datasets for training damage assessment models to date \cite{gupta2019creating, rahnemoonfar2021floodnet, rahnemoonfar2022rescuenet, zhu2021msnet, cheng2021dorianet, IdaBDDataset, manzini2024crasar, haitiBRDDataset, liuLargeScale2019, chen2018benchmark, fujita2017damage, Pi2020, IanBD, chen2025bright} incorporate the assumption that aerial imagery derived labels are sufficient as they follow the same general process of first labeling input aerial images, then training and testing damage assessment models on labels derived from aerial images instead of labels from ground level inspections or other sources. The previous datasets from which models could be trained differ in source and size, with the largest drone and satellite imagery datasets containing 21,716 buildings \cite{manzini2024crasar} and 850,736 buildings \cite{gupta2019creating} respectively, and the labeling schema, which may be ad hoc as in \cite{Pi2020, cheng2021dorianet, zhu2021msnet, rahnemoonfar2021floodnet, fujita2017damage}. A corollary assumption about aerial imagery is that multiple views will be consistent and that satellite and drone imagery will independently produce the same damage labels for a building. Multi-view equivalence means that disaster management agencies could choose to use whichever source is available or more cost-effective, since the model performance will be approximately equivalent.

However, if these assumptions are faulty, they could lead to incorrect interpretations with serious societal and ethical consequences. A 2024 National Academies of Sciences Engineering and Medicine study \cite{NASGCC:24} reported negative consequences from drone imagery being mislabeled by humans in the aftermath of Hurricane Laura in Lake Charles, Louisiana. Drone imagery was presumed to be a satisfactory substitute for physical ground-level inspections of buildings. Unfortunately, this mislabeled damage caused repairable houses to be demolished at great cost to the individuals and the agencies. The study also states that mislabeling led to those agencies rejecting the use of virtual inspections for future events. Hurricane Laura serves as a warning for machine learning researchers that relying only on aerial imagery-derived labels of damage could result in costly mistakes and erode trust in the technology.

This paper reports on a first step to quantitatively address whether aerial imagery-derived labels can accurately represent disaster scenes and be used to support rapid, trustworthy, automated damage assessment. Although this paper does not have access to ground-level labels, it explores the assumptions by performing a data audit comparing damage labels for a total of 15,814 buildings spanning three major hurricanes (Ian, Michael, Harvey) where spatially and temporally coincident satellite and drone imagery was available. The hurricanes resulted in varying levels of damage due to storm surge, flooding, and wind damage. Taken together, they offer a comprehensive dataset of damage. Images were labeled with the Joint Damage Scale (JDS) schema \cite{gupta2019creating}, which has been widely used \cite{gupta2019creating, IdaBDDataset, IanBD, haitiBRDDataset, manzini2024crasar}.

\begin{table*}[]
\begin{tabular}{|c|c|c|c|}
\hline
\textbf{Sensing Platform}  & \textbf{Dataset Name}                        & \textbf{Release Year} & \textbf{Label Source} \\ \hline
\multirow{6}{*}{Drone}     & CRASAR-U-DROIDs \cite{manzini2024crasar}     & 2024                  & Imagery-Derived       \\ \cline{2-4} 
                           & RescueNet \cite{rahnemoonfar2022rescuenet}   & 2023                  & Imagery-Derived       \\ \cline{2-4} 
                           & ISBDA \cite{zhu2021msnet}                    & 2021                  & Imagery-Derived       \\ \cline{2-4} 
                           & DoriaNET \cite{cheng2021dorianet}            & 2021                  & Imagery-Derived       \\ \cline{2-4} 
                           & Volan v.2018 (Drone) \cite{Pi2020}           & 2020                  & Imagery-Derived       \\ \cline{2-4} 
                           & FloodNet \cite{rahnemoonfar2021floodnet}     & 2020                  & Imagery-Derived       \\ \hline
\multirow{5}{*}{Satellite} & BRIGHT \cite{chen2025bright}                 & 2025                  & Imagery-Derived       \\ \cline{2-4} 
                           & HaitiBRD  \cite{haitiBRDDataset}             & 2023                  & Imagery-Derived       \\ \cline{2-4} 
                           & Ida-BD \cite{IdaBDDataset}                   & 2022                  & Imagery-Derived       \\ \cline{2-4} 
                           & xBD \cite{gupta2019creating}                 & 2018                  & Imagery-Derived       \\ \cline{2-4} 
                           & ABCD \cite{fujita2017damage}                 & 2017                  & Imagery-Derived       \\ \hline
\end{tabular}
\caption{Aerial Imagery Datasets for Damage Assessments. Datasets are grouped by sensing platform, and ordered by most to least recent release year. All datasets for building damage assessment from aerial imagery rely on imagery-derived labels, and none are aligned with ground-level labels.}
    \label{tab:datasets}
\end{table*}

The audit shows that not considering ground-level labels and the existing approach of learning labels from only aerial imagery may lead to systemic misclassification and, as shown with Hurricane Laura, follow-on societal harm and loss of trust. The audit finds that damage labels derived from satellite imagery and drone imagery have significant label disagreement, with 29.02\% of the buildings having different damage labels between satellite imagery and drone imagery. Examples of this disagreement are shown in Figure \ref{fig:suas-vs-satellite}. One possible interpretation of the label disagreements is that assessments from drone imagery are more accurate than satellite images because drone imagery has a higher resolution (2-5cm/pixel for drone vs 30-80cm/pixel for satellite). But, there is the possibility that neither label is correct, especially since damage from flooding or storm surge may not be fully visible from overhead. Thus, it cannot be determined if one source is intrinsically better without examining the ground-level labels. The audit also refutes the multi-view assumption that the two sources can be used interchangeably. The satellite-derived labels represent a different distribution of damage labels from drone-derived imagery. Models trained and evaluated on these two sources of aerial imagery will produce different labels, but without ground-level labels, the correct label cannot be determined.

The remainder of the paper is organized as follows. Section \ref{sec:related-work} provides background on the operational use of aerial imagery for damage assessment and details the existing datasets and work in labeling. Section \ref{sec:approach} details the satellite and drone imagery from Hurricanes Ian, Michael, and Harvey used for the audit, the Joint Damage Scale \cite{gupta2019creating} schema, and the labeling workflow, including a two-stage review process. The analysis is presented in Section \ref{sec:analysis}, describing how specific satellite views of a building were selected (Section \ref{subsec:view-select-strategies}), then applying Chi-Squared tests and z-tests revealing that satellite views significantly under-report damage compared to drone views (Section \ref{subsec:drone-satellite-class-balances}). The analysis also computes the overall label disagreement (Section \ref{subsec:drone-satellite-label-disagreement}). The discussion of the findings in Section \ref{sec:discussion} notes the limitations of this audit (Section \ref{subsec:limitations}), but also the value of the analysis in terms of considering 19.05 times more buildings than previous work with the same labeling schema (Section \ref{subsec:analysis_properties}). The discussion goes further by summarizing the implications of automated building damage assessment (Section \ref{subsec:implications_cvml}) and calling out the potential risks and harms to disaster operations from using models that learn only from aerial imagery (Section \ref{subsec:risks_harms_ops}). Section \ref{sec:conclusion} concludes with four recommendations for improving the reliability of CV/ML models and transparency to decision-makers as well as outlines ongoing and future work.

\section{Related Work}
\label{sec:related-work}

Prior literature shows no evidence of qualitative or quantitative comparisons between satellite- and drone-derived labels or validated labels sourced from imagery and any known ground-level labels. Further, there are no known datasets that could be used, or combined, to perform such a comparison. Instead, aerial imagery datasets for damage assessments rely on imagery-derived labels with an unknown relationship with ground-level labels. This presents a knowledge gap in understanding the difference between imagery sources and their alignment with the actual conditions as represented by ground-level labels. This section will discuss the use of satellite and drone imagery (Section \ref{subsec:comparison-satellite-drone-imagery}), present the current state of aerial post-disaster imagery datasets (Section \ref{subsec:datasets}), and summarize the approach and results of two prior efforts comparing imagery-derived labels to ground-level labels (Section \ref{subsec:comparison-with-ground-truth-labels}).

\subsection{Comparison of Satellite and Drone Imagery}
\label{subsec:comparison-satellite-drone-imagery}

Although satellite and drone imagery have become commonplace in disaster response and recovery \cite{LozanoIJDRR:23}, no studies appear to quantitatively compare their use for damage assessment, either from manual inspection of the imagery or by automated classification. If both sources were available at equal cost, following \cite{botzen:19}, a local agency might prefer drone imagery because of its intrinsic higher resolution or satellite imagery due to its greater coverage of the affected area. In either case, CV/ML techniques would be needed to automate rapid damage assessment given both satellites and drones contribute unmanageably high volumes of imagery \cite{AnkerAOR:19, garcia2021computer}. 

However, it is unlikely that both sensor sources would be simultaneously available because of the notable technological and institutional differences that can delay the timely delivery of imagery \cite{LozanoIJDRR:23, GFDRR:19, manzini2023wireless}. Without quantifiable trade-offs, local agencies would lack a firm basis for choice. 
From a technological perspective, there are five major differences: air space regulations, weather, spatial coverage, spatial resolution, and temporal resolution. The air space for drones may be restricted by aviation regulations whereas satellites do not need permissions. Both can be impacted by weather, though drones can often collect imagery flying under thick clouds that might block satellite electro-optical cameras. Satellites typically cover much larger areas, such as counties versus a neighborhood covered by a drone. Satellites and drones have different spatial resolutions, with common satellite imagery having 30cm/pixel and drones at 3cm/pixel. The two sources also vary in how quickly they can be deployed to an area. As shown in \cite{manzini2023quantitative, fernandes2019quantitative, fernandes2018quantitative}, drones are typically deployed the day after the disaster whereas it might take days to reposition or redirect a satellite. Lag time in data availability is a major institutional difference between satellites and drones. Satellites are strategic assets that do not belong to a local agency whereas drones are local and imminently taskable. Satellite data may be expensive, if from a commercial source, or have restricted sharability if from a military or intelligence source. As a local asset, drones offer greater flexibility in tasking and the imagery is immediately available to local agencies flying the drones, even in rural areas without internet. If stakeholders are wholly reliant on wireless connectivity, drone imagery may be delayed as it often needs to be physically transported out of the disaster scene \cite{manzini2023wireless} while satellites transmit their imagery directly to centralized data stores making imagery readily available for transmission. 

\subsection{Datasets}
\label{subsec:datasets}

% TODO: Consider rephrasing imagery-derived labels to something that makes it clear that this imagery is from aerial imagery
No datasets for aerial imagery-based damage assessment consider label agreement across views or with ground-level labels, and all rely on imagery-derived labels. Single views and lack of ground-level labels limit the understanding of the agreement between imagery-derived multi-view damage labels and the actual condition. 

11 aerial imagery post-disaster datasets \cite{gupta2019creating, rahnemoonfar2021floodnet, rahnemoonfar2022rescuenet, zhu2021msnet, cheng2021dorianet, IdaBDDataset, manzini2024crasar, haitiBRDDataset, fujita2017damage, Pi2020, chen2025bright} intended for training automated damage assessments systems, for either drone or satellite imagery, were identified from prior literature. These datasets are shown in Table \ref{tab:datasets}, consisting of five drone \cite{rahnemoonfar2021floodnet, rahnemoonfar2022rescuenet, cheng2021dorianet, manzini2024crasar, zhu2021msnet} and 
%three crewed aircraft \cite{IanBD, liuLargeScale2019, chen2018benchmark}, 
five satellite \cite{gupta2019creating, IdaBDDataset, fujita2017damage, haitiBRDDataset, chen2025bright} datasets, and one \cite{Pi2020} which features both drone and crewed aircraft imagery. 
% It is worth noting that the \citeauthor{chen2018benchmark} dataset\cite{chen2018benchmark} includes model-based labels, from a non-imagery-based flood inundation model, and reports including imagery-derived labels; however, these imagery-derived labels do not appear to be available. 
All 11 datasets' damage labels are sourced solely from imagery.

% - Removed as chen 2018 is manned 
% The \citeauthor{chen2018benchmark} dataset\cite{chen2018benchmark} is the only dataset that applies more than one damage label for the same building. However, these labels do not use a consistent damage schema,  and while this work reports providing crewed aircraft imagery-derived damage labels and synthetic model-derived labels, the imagery-derived labels are no longer available. As a result, it is not possible to use this data for an audit like this.

The Volan v.2018 dataset \cite{Pi2020} is the only dataset that labels damage from multiple sources, drones and crewed aircraft, using the same damage scale. However, this dataset does not visually appear to contain coincident imagery of the same scenes from the two different sources, and it is not georeferenced. The lack of georeferencing means that even if the dataset did contain coincident imagery of the same scenes, it would require substantial effort to determine any coincident labels. Practically, this means that this data cannot be used for an audit like this.

% Except for \cite{chen2018benchmark}, discussed above, 
All datasets provide a single view of a building and a single damage label derived from aerial imagery. Further, the differing damage schemas present throughout these datasets make it inappropriate to compare the damage labels across datasets for what little coincident data may exist. 
This represents a gap in the literature as there does not appear to be any publicly available data that could be used to validate the degree to which the aerial imagery-derived labels in these datasets align with the operational objectives the datasets attempt to address.

%Without aligned damage schemas and a consistent mechanism for identifying coincident buildings, any labels for building damage across these datasets cannot be faithfully compared. 

\subsection{Comparison with Ground-Level Labels}
\label{subsec:comparison-with-ground-truth-labels}
There are no known efforts in the literature that attempt to determine the extent to which \textit{human-sourced} imagery-derived labels agree with one another or ground-level assessments of building damage; however, two instances of comparing \textit{automated} imagery-derived labels with ground-level labels do exist \cite{robinson2023rapid, sadek2021engineering}. Both approaches are limited by differing damage schemas between the ground-level and aerial imagery, potentially misaligned locations of buildings and the locations of ground-level assessment labels, and lower quantities of data than this work.

In \citep{robinson2023rapid}, the authors compare 830 ground-level labels to electro-optical imagery-derived labels produced by a CV/ML model. The comparison was performed by reducing both labels to a collection of binary labels representing ``damage" and ``no damage" classes. Then the imagery-derived labels were evaluated using binary precision, recall, and average precision metrics. The authors compare two different strategies to convert the ground-level labels to binary labels and find that the average precision of the model-based imagery-derived labels compared to the ground-level labels is 0.81 or 0.86, depending on the label conversion strategy. The authors align the ground-level damage assessments with known building locations by mapping a single ground-level label to any buildings within a 20-meter radius of the ground-level label location. This results in some ground-level labels not having an associated known building location, and the authors report successfully mapping 830 out of 909 ground-level labels \cite{robinson2023rapid}. To the best knowledge of the authors, this work represents the largest comparison of imagery-derived labels to ground-level damage assessments.

In \citep{sadek2021engineering}, the authors compare 172 ground-level building damage labels to synthetic aperture radar (SAR) imagery-derived labels produced by the ``Advanced Rapid Imaging and Analysis (ARIA) team at the California Institute of Technology (Caltech) and the NASA—Jet Propulsion Laboratory (JPL)" from the area impacted by the 2020 Beirut explosion. The 172 ground-level labels were derived from in-person inspections of buildings and individual buildings were categorized into ``no damage," ``light damage," ``moderate damage," ``heavy damage," "partial structural collapse," and ``full structural collapse." The SAR imagery-derived labels were provided at the pixel level with a reported ground sampling distance of ``about 10 by 10 meters" per pixel. The pixel-level labels were numeric and ranged between 0 and 1.  It is unclear how the authors converted from these pixel-level labels to the individual buildings that were inspected. The comparison was performed by measuring the correlation between the damage labels produced via ground-level inspections and the SAR imagery-derived labels. The authors find that a correlation does exist and that the SAR imagery-derived labels possess ``an ability [...] to distinguish among damage levels" collected via ground-level inspection.

\section{Approach}
\label{sec:approach}
%Priya: Perhaps this shouldn't be the topic sentence...
\begin{table*}[]
\begin{tabular}{|c|c|c|c|}
\hline
\textbf{Disaster Event} & \textbf{\# Drone Orthomosaics} & \textbf{\# Satellite Orthomosaics} & \textbf{\# Coincident Buildings} \\ \hline
Hurricane Ian     & 25                      & 62                          & 14,268                            \\ \hline
Hurricane Michael & 2                       & 4                           & 1,142                             \\ \hline
Hurricane Harvey  & 3                       & 5                           & 404                              \\ \hline
\textbf{Total}    & \textbf{30}             & \textbf{71}                 & \textbf{15,814}                   \\ \hline
\end{tabular}
\captionof{table}{Views and Coincident Buildings per Disaster Event. Ordered from most to least number of coincident buildings.} % TODO: Update number of coincident buildings
    \label{tab:tab-views}
\end{table*}

\begin{figure}
    \centering
    \includegraphics[width=0.45\textwidth]{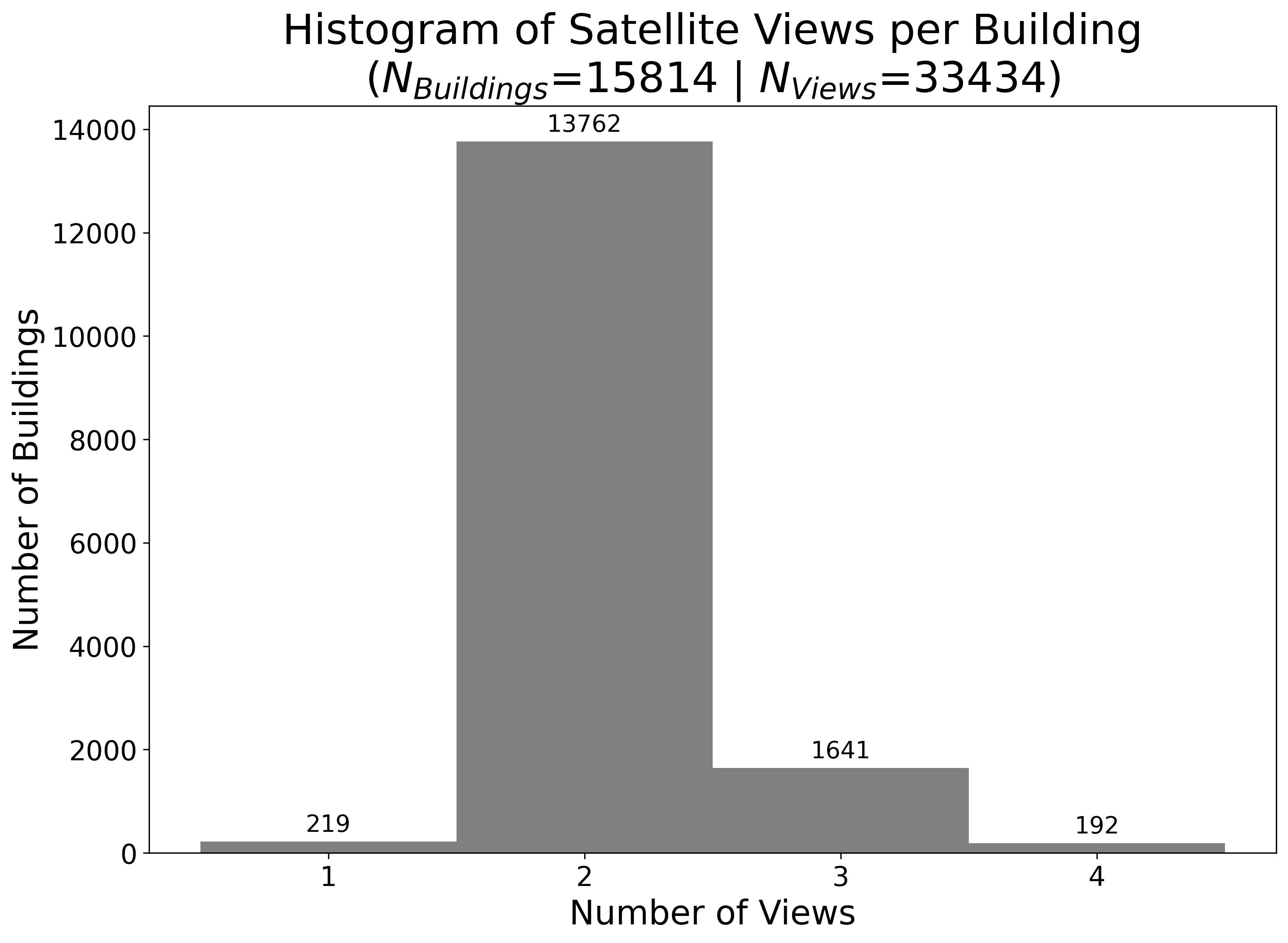}
    \caption{Satellite Views per Building. Each building averages 2.11 satellite views.}
    \label{fig:bar-chart-views}
\end{figure}

This audit directly compares both drone- and satellite-derived damage labels for the same buildings using the Joint Damage Scale (JDS) label schema\cite{gupta2019creating} across the three hurricanes (Ian, Michael, and Harvey), examples of which are shown in Figure \ref{fig:suas-vs-satellite}. The data audited consists of drone-derived labels, sourced from the CRASAR-U-DRIODs dataset \cite{manzini2024crasar}, and satellite-derived labels, labeled by a pool of annotators for satellite imagery sourced from Maxar Open Data Portal \cite{maxar15}, for 15,814 coincident buildings.
%All damage labels use the same damage schema, JDS, to ensure alignment with the operational use case and modality of data. -> Tom: I think it is fine to delete.
This section will detail the data audited (Section \ref{subsec:data}), the damage label schema (Section \ref{subsec:label-schema}), and the labeling and review workflow for satellite-derived damage labels (Section \ref{subsec:labeling-workflow}). 

\subsection{Data}
\label{subsec:data}

% Priya: Is it okay if we say views instead of orthomosaics, to avoid having to define what an orthomosaic is? 
% Tom: I think we need to define orthomosaics in some way since we need to have some way of describing how the satellite imagery was bounded. Orthomosaics boundaries, etc...
As shown in Table \ref{tab:tab-views}, the audited data consists of 15,814 buildings from 30 post-disaster drone orthomosaics, from the CRASAR-U-DRIODs dataset\cite{manzini2024crasar}, and 71 post-disaster satellite orthomosaics, from Maxar Open Data Portal\cite{maxar15}, from Hurricanes Ian, Michael, and Harvey. In total, this represents 33,434 total views of the 15,814 buildings, meaning that each building was captured in satellite imagery and thus labeled an average of 2.11 times. The distribution of satellite view counts per building is shown in Figure \ref{fig:bar-chart-views}.
Pre-disaster views are not considered in this work. Crucial to this audit, the 15,814 coincident buildings refer to the same building locations present in both drone and satellite views. 
The majority of buildings are from Hurricane Ian (14,268 coincident buildings), followed by Hurricane Michael (1,142 coincident buildings) and Hurricane Harvey (404 coincident buildings).

All satellite imagery considered within this analysis has spatial and disaster event overlap with the drone imagery from the CRASAR-U-DRIODs dataset\cite{manzini2024crasar}, is captured post-disaster and, per established practices for automated building damage assessment, is less than 80cm/px \cite{gupta2019creating}. Also, it is captured between 5 days before or 2 days after the drone imagery. These date ranges do not represent specific criteria, but the totality of imagery available on the Maxar Open Data Portal for the drone imagery is considered in this analysis.

One collection of drone imagery contained in CRASAR-U-DROIDs did contain coincident satellite imagery in the Maxar Open Data Portal \cite{maxar15} and was excluded from this analysis\footnote{090403-Lancaster-Canyon-Gate.geo.tif} because it contained flooding conditions that were substantially different from the drone imagery. This collection of imagery was excluded due to this audit's intention to measure label disagreement across views where temporal factors, such as changing flood waters, are not a factor. 

For the imagery considered, there is no evidence that drones were directed to areas where satellite imagery was unavailable or insufficient for damage assessments, and vice versa. 
After consultation with the authors of CRASAR-U-DROIDs \cite{manzini2024crasar}, and referencing prior work \cite{manzini2023quantitative, fernandes2018quantitative, fernandes2019quantitative, manzini2023wireless}, it appears that the drone imagery collection was unrelated to the satellite imagery or was performed without knowledge of where satellite imagery had been or would be collected. 
To the authors' best knowledge, drone and satellite imagery were collected independently.

\begin{figure}
    \centering
    \includegraphics[width=0.9\linewidth]{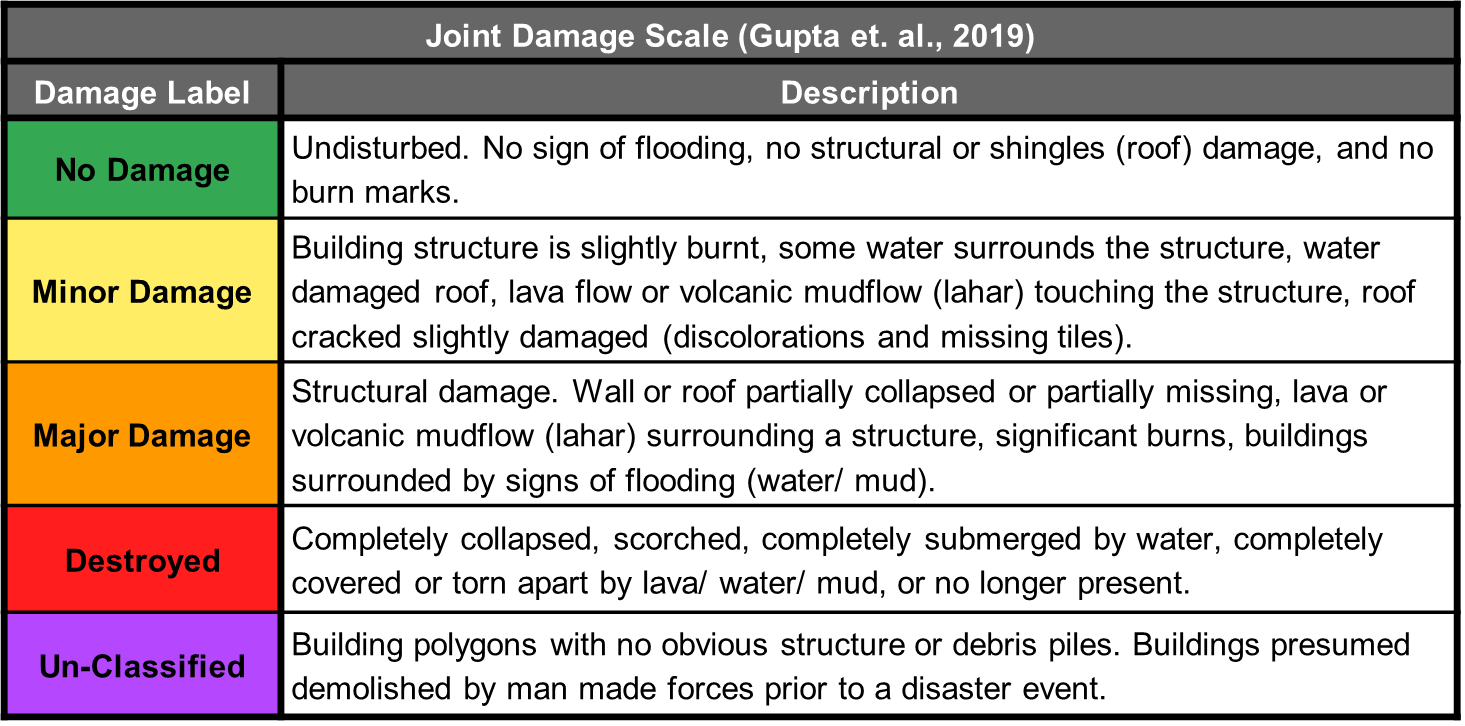}
    \captionof{table}{Joint Damage Scale Schema Adapted from \cite{gupta2019creating} detailing five damage classes and their respective descriptions.}
    \label{fig:label-schema}
\end{figure}

\subsection{Damage Label Schema}
\label{subsec:label-schema}

All buildings in the satellite imagery were labeled via the Joint Damage Scale (JDS) \cite{gupta2019creating}, a five-class damage schema and the only building damage schema tailored for remote sensing. This damage schema is described in Table \ref{fig:label-schema}. This was an intentional choice as it is the damage schema used in the CRASAR-U-DROIDs dataset, and thus, using this schema would allow for labels to be directly compared across views, enabling this work to overcome the limitations within the literature as discussed in Sections \ref{subsec:datasets} and \ref{subsec:comparison-with-ground-truth-labels}.

As described in Table \ref{fig:label-schema}, the JDS consists of five classes for building damage assessment: ``no damage", ``minor damage", ``major damage", ``destroyed", and ``un-classified." A ``no damage" label refers to an undistributed building that has no signs of damage. A ``minor damage" label refers to a building with slight indications of damage but no structural damage. A ``major damage" label refers to a building that has endured structural damage. A ``destroyed" label refers to a building that has been destroyed and no longer has a standing structure. Lastly, an ``un-classified" label refers to a building polygon, the building footprint location, that does not correspond to a building. 

% -- Removed discussion on obscured
% One gap in the JDS is that of obscured buildings. Object occlusion in satellite imagery is a well-known and studied phenomenon \cite{chen2007occlusion, su2016shadow, senlet2012segmentation, jeppesen2019cloud} and the analysis presented in this paper found buildings obscured by clouds, trees, other buildings, and lens flares. In these cases, it was not feasible to provide accurate building damage labels that could be faithfully compared to drone-derived labels. To handle this case, buildings that could not be observed due to occlusion were labeled as ``obscured."

%To overcome this limitation, the JDS was utilized as it i JDS is catered to aerial imagery, making it an appropriate schema for the drone- and satellite-derived labels audited. 

\begin{figure*}
    \centering
    \includegraphics[width=0.95\textwidth]{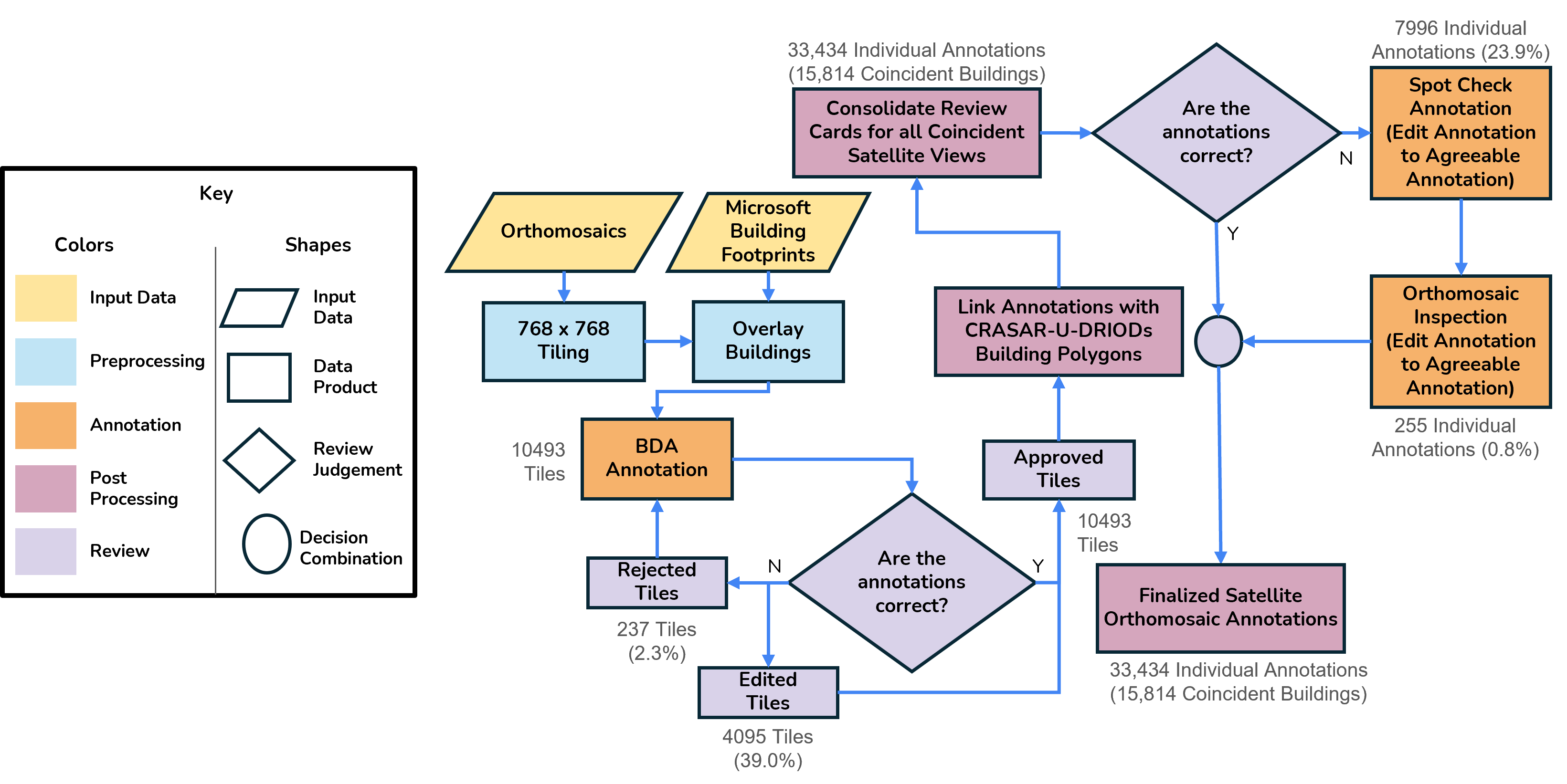}
    \caption{Labeling workflow for satellite-derived damage labels. Each satellite-derived damage label was generated from annotations for building polygons, according to the Joint Damage Scale (JDS), and reviewed via a 2-stage review process.}
    \label{fig:labeling-workflow}
\end{figure*}

%% TODO Update the below line to once the we do not have be anonymous
This work was performed with consultation and guidance from the authors of the CRASAR-U-DROIDS dataset \cite{manzini2024crasar} and the xBD dataset \cite{gupta2019creating}. 
This was done to ensure the JDS was applied consistently and faithfully compared to the drone-derived labels present in the CRASAR-U-DROIDs dataset. Additionally, this ensures that any label disagreement observed between the drone- and satellite-derived labels is a result of actual changes in information present in the imagery as opposed to misapplications of the JDS.

\subsection{Labeling Workflow}
\label{subsec:labeling-workflow}

As shown in Figure \ref{fig:labeling-workflow}, 33,434 images, sourced from Maxar Open Data Portal\cite{maxar15}, containing buildings coincident with the drone imagery were annotated by 83 annotators from a pool of 157 annotators according to the JDS and reviewed via a 2-stage review process. This workflow consisted of five components: input data, preprocessing, annotation, post-processing, and review. 

The input data consisted of 71 satellite orthomosaics, aerial maps that are geospatially aligned, and building footprints. The satellite orthomosaics were sourced from the Maxar Open Data Portal\cite{maxar15}. The building footprints were sourced from the same Microsoft Building Footprints \cite{MicrosoftBuildingFootprints} as the CRASAR-U-DROIDs dataset \cite{manzini2024crasar}.

Preprocessing consisted of tiling the orthomosaics and overlaying building footprints. Each orthomosaic was tiled into 768x768 image tiles and then overlaid with building footprints located within them. This resulted in 10,493 image tiles that were uploaded to Labelbox \cite{Labelbox2024}, a web-based data labeling tool. 

83 annotators participated in the generation of these labels. Within Labelbox, each annotator provided labels for the 10,493 image tiles, resulting in 33,434 individual annotations for the 15,814 buildings. Later on, three committee members of reviewers made 7,996 corrections to these annotations to ensure the satellite-derived labels agreed with the JDS.

Postprocessing linked annotations with the same building polygons from the CRASAR-U-DRIODs dataset, consolidated coincident satellite views of the same buildings for review, and finalized satellite orthomosaic annotations. Since the building polygons were the same as those in CRASAR-U-DROIDs, linking annotations to CRASAR-U-DROIDs polygons was straightforward but required discarding or correcting instances where annotators created or edited polygons.
Once the annotations were linked to the CRASAR-U-DRIODs building polygons, all 33,434 satellite views were grouped by building so they could be reviewed together at a building level. After all annotations and the 2-stage reviews, building polygons were finalized, resulting in 15,814 coincident buildings with 33,434 finalized damage labels. 

%While we used the term review card internally, lets not discuss it here as I dont think it is worthwhile to define it. Lets instead define this process generically. If we want, we can put an example of a few cards in the supplementary data/info - Tom
The 2-stage review process consisted of a tile-level review and a secondary building-level review. At the tile-level review, a single reviewer would inspect an image tile and either reject or approve the tile. As a result of this process, 237 (2.3\%)
image tiles were rejected outright and returned to annotators, and 4,095 (39.0\%) image tiles were edited by reviewers in some way. In all, 4,332 (41.3\%) image tiles were either rejected or had their labels edited by reviewers. Following this process, all 10,493 image tiles were approved. After that, a committee of at least two of the authors inspected the labels for each view of a single building simultaneously and updated the labels for each view where the committee deemed appropriate. During the building-level reviews, the committee discussed each annotation for all 33,434 views of all 15,814 buildings. This building-level review resulted in the change of 7,996 annotations (23.9\%). Once the spot checks were completed, a final inspection of the complete orthomosaics was performed, and 255 individual labels were updated (0.8\%).

The motivation for this additional building-level committee review step was three-fold. First, it replicates the second stage committee review that was performed in the creation of the CRASAR-U-DROIDs labels\cite{manzini2024crasar}. Second, it provided an additional check to ensure that the JDS was applied consistently. Finally, it allowed the committee a chance to correct any misunderstandings or errors made by annotators or in one of the earlier reviews. At no point during the annotation or review process were the annotators or reviewers presented with coincident drone imagery or labels. This ensured that no satellite labels could be influenced by drone imagery labels.

\begin{figure*}
    \centering
    \includegraphics[width=0.95\textwidth]{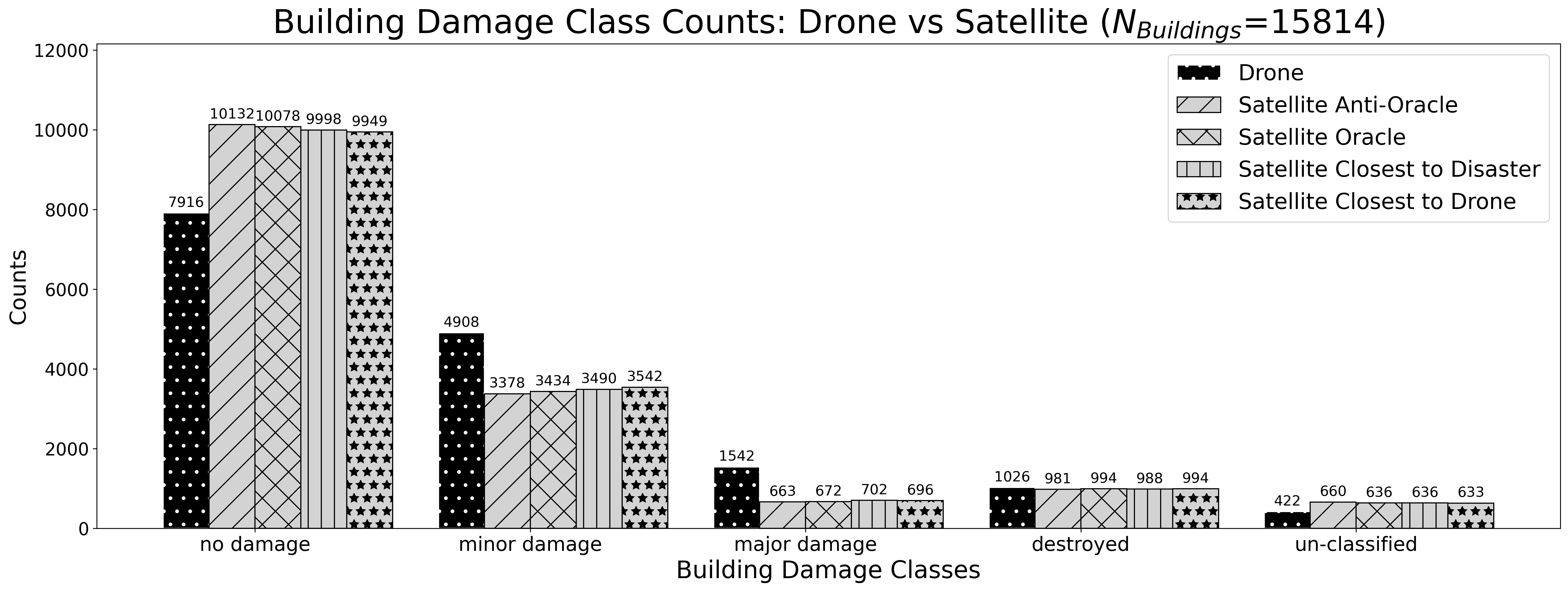}
    \caption{Building Damage Class Counts: Drone vs Satellite Class Balances. This figure presents the labels for 15,814 coincident buildings across drone and satellite views. Each bar represents either a drone-derived damage label or a satellite-derived damage label from each of the four view selection strategies.}
    \label{fig:label-counts}
\end{figure*}

\section{Analysis}
\label{sec:analysis}

Three analyses conducted analyzed 15,814 coincident buildings across drone and satellite views, finding that satellite-derived damage labels significantly under-report damage compared to drone-derived damage labels, and satellite- and drone-derived damage label distributions differ significantly. The three analyses conducted measured damage label distributions of satellite- and drone-derived labels, label disagreement rates across coincident satellite imagery, and factors that influence these distributions. The remainder of this section further details the view selection strategies used for the analyses (Section \ref{subsec:view-select-strategies}), and the three analyses: comparison of satellite- and drone-derived damage labels (Section \ref{subsec:drone-satellite-class-balances}), comparison of label disagreement between satellite- and drone-derived damage labels (Section \ref{subsec:drone-satellite-label-disagreement}), and comparison of label disagreement across coincident satellite imagery (Section \ref{subsec:paralle-satellite-label-disagreement}).

%, comparison of changes in view properties and damage label changes (Section \ref{subsec:satellite-view-properties}).

\subsection{View Selection Strategies for Analysis}
\label{subsec:view-select-strategies}
Four view selection strategies were used to measure differences between drone- and satellite-derived labels, selecting a single view from the multiple satellite views included in this analysis. The view selection strategies are as follows.

\begin{itemize}
    \item \textit{Satellite Closest to Drone}: The closest satellite view of the coincident building to the drone view capture date. This strategy represents the closest comparison to the conditions of buildings at the time of the drone imagery.
    \item \textit{Satellite Oracle}: The satellite view of the coincident building whose damage label agrees with the drone view damage label. This strategy represents a lower bound for label disagreement between drone- and satellite-derived labels by sharing label information between the two groups to decrease label disagreement.
    \item \textit{Satellite Anti-Oracle}: The satellite view of the coincident building whose damage label does not agree with the drone view damage label. This strategy represents an upper bound for label disagreement between drone- and satellite-derived labels by sharing label information between the two groups to increase label disagreement.
    \item \textit{Satellite Closest to Disaster}: The closest post-disaster satellite view of the coincident building to the disaster date. This strategy selects the earliest known damage label to capture buildings before any repairs or cleanup.
\end{itemize}

\subsection{Drone- and Satellite-Derived Label Class Balances}
\label{subsec:drone-satellite-class-balances}

As shown in Figure \ref{fig:label-counts}, the satellite views' damage label counts differ significantly from the drone's for each damage label, except the ``destroyed" label. At a damage label level, a z-test revealed that satellite views significantly under-report damage compared to drone views for ``no damage" (p<1.2x10\textsuperscript{-117}), ``minor damage" (p<1.7x10\textsuperscript{-67}), and ``major damage" (p<1.4x10\textsuperscript{-75}) labels and report equivalent counts compared to the drone views for the "destroyed" label (p>0.30).

At the individual damage label level, not only are the label disagreements between drone- and satellite-derived labels significant, but they are also large in magnitude and with insignificant differences depending on the view-selection strategy (p>0.14).
When compared to the drone-derived labels, satellite-derived labels appear to over-report the ``no damage" label by 20.8\%-21.9\%, under-report the ``minor damage" label by 38.6\%-45.3\%, and under-report ``major damage" by 132.6\%-119.7\%.
The ``destroyed" label again represents an exception, as label disagreement between drone- and satellite-derived labels is not significant and varies between 3.2\%-4.6\%.
Finally, the ``un-classified" label appears to be over-reported in satellite-derived labels compared to drone-derived labels by 33.3\%-36.1\%.

\subsection{Drone- and Satellite-Derived Label Disagreement}
\label{subsec:drone-satellite-label-disagreement}
Considering these distributions more generally, instead of at an individual label level, the proportion of labels that disagree between satellite- and drone-derived labels is between 0.29 and 0.36, depending on which view selection strategy was used. %This means that %and the different satellite view selection strategies, discussed in Section \ref{subsec:satellite-view-properties}, represent equivalent distributions when compared to one another.
The disagreement rate between the satellite oracle and drone is 0.29. ``Satellite closet to drone and drone" the disagreement rate is 0.32. ``Satellite closest to disaster and drone" disagreement rate is 0.33. The disagreement rate between ``satellite anti-oracle and drone" is 0.36. 

These disagreements are statistically significant, and drone-derived labels represent a significantly different distribution than those generated by all different satellite view selection strategies based on a Chi-Squared test (p<5.1x10\textsuperscript{-175}). It is worth noting that when comparing the different satellite view selection strategies to one another, there are no significant differences based on a Chi-Squared test (p>0.14).

%The differences in proportions of ``no damage," ``minor damage," ``major damage," and ``unclassified" labels between drone-derived labels and all satellite view selection strategies are statistically significant based on a z-test with $\alpha=0.001$. It should be noted that ``destroyed" building labels are an exception, and the differences between drone- and satellite-derived labels are not statistically significant (p>0.79).

\begin{figure}
    \centering
    \includegraphics[width=0.45\textwidth]{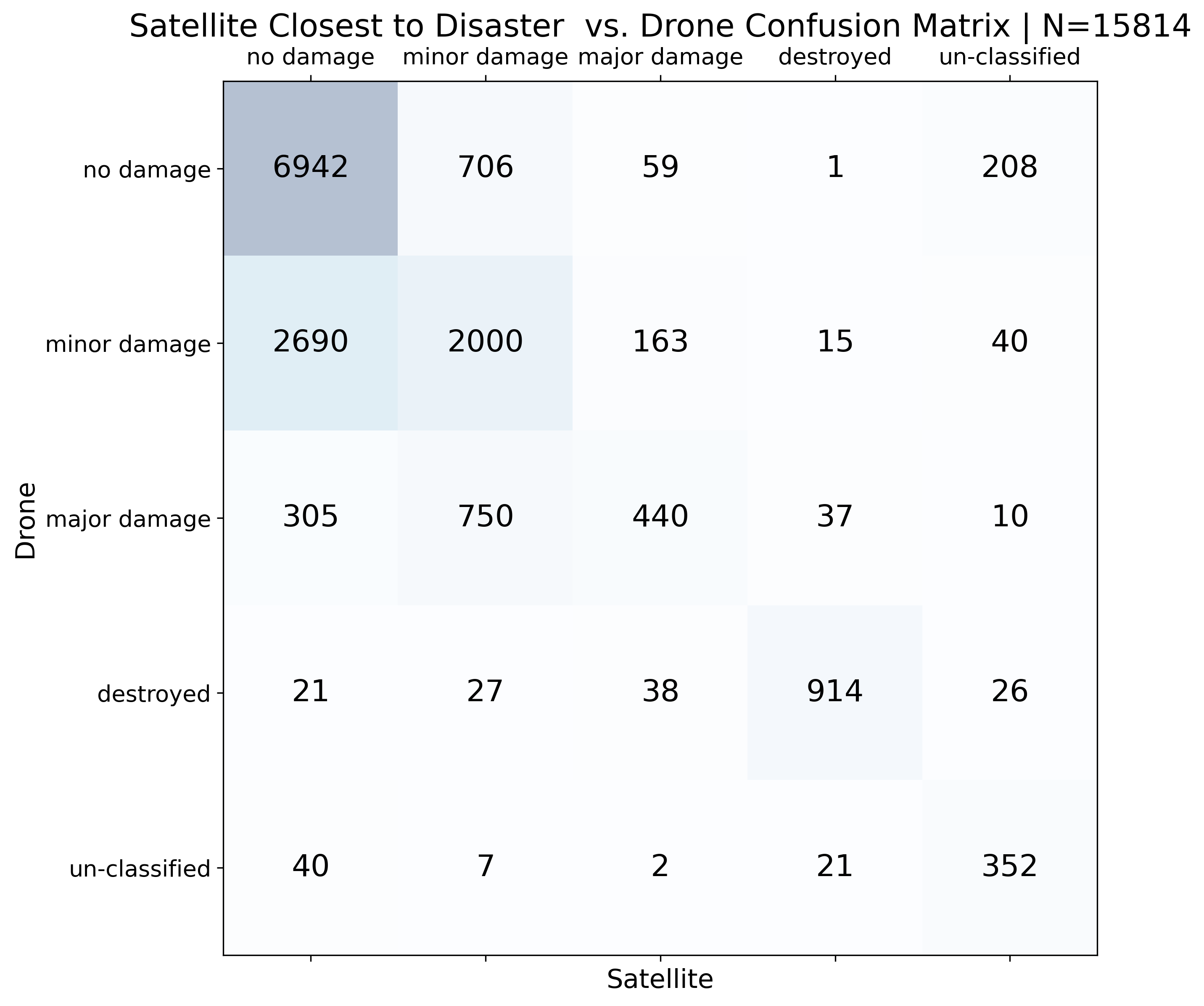}
    \caption{Building Damage Class Drone vs Satellite View Closest to Drone Confusion Matrix. The y-axis is the drone-derived damage labels, and the x-axis is the satellite-derived damage labels for the 15,814 Coincident Buildings.} 
    \label{fig:transition-matrix}
\end{figure}

Figure \ref{fig:transition-matrix} shows the explicit mapping between drone- and satellite-derived labels for the ``satellite closest to drone" strategy as it is believed that this strategy represents the most honest presentation of data as it would appear in real-world conditions. Based on this figure, satellite-derived labels appear to under-report damage when compared to drone-derived labels. In some cases, this effect is substantial. Of the buildings that were labeled as ``minor damage" in the drone imagery, 1.32 times more buildings were labeled as ``no damage" compared to being labeled as ``minor damage." This effect is more pronounced when considering the buildings that were labeled as ``major damage" in the drone imagery, with 2.38 times more buildings being labeled as either ``no damage" or ``minor damage".

\subsection{Coincident Satellite-Derived Label Disagreement}
\label{subsec:paralle-satellite-label-disagreement}
A natural question when comparing multiple views of a scene from differing sensors is whether alternate dynamics beyond label source (drone vs satellite) are influencing changes in labels between the different distributions. To quantify this, the label disagreement rates between labels derived from differing coincident satellite views were computed.

To measure the disagreement rates between coincident satellite views, all pairs of coincident satellite imagery for each building were selected, their labels were compared, and their respective disagreement rates were computed.  This resulted in 16,261 satellite view pairs to be considered.

%Satellite view pairs in which at least one view of a building was labeled as obscured were not considered in this analysis.

The disagreement rate across the 16,261 coincident satellite view pairs is 0.074. This represents a disagreement rate of 3.92 times less than the minimum disagreement rate observed between ``satellite oracle and drone", described in Section \ref{subsec:drone-satellite-label-disagreement}. The scale of label disagreement between satellite view pairs and coincident drone and satellite imagery indicates that the label disagreement observed in coincident satellite and drone imagery cannot reasonably be explained by label variations within the satellite imagery alone.

%%% - Removed as it does not move the claims - %%%
% \subsection{Satellite View Properties Indicative of Label Disagreement}
% \label{subsec:satellite-view-properties}

% From a chi-squared test? (TODO: Which test? and state alpha values for significance), the differences between the satellite oracle and satellite anti-oracle are not significant (p=0.0515). 

% Changes in view properties, date of capture, sun elevation, sun azimuth, and view azimuth, are indicative of label disagreement between parallel satellite views. 89.2\% (30596/34305) of satellite building views have metadata for six view properties: date of capture, off nadir, sun elevation, view azimuth, incident angle, and sun azimuth. % TODO: Maybe define what these are.
% To determine whether a change within these view properties would be indicative of label disagreement, the satellite view pair, described in Section \ref{subsec:paralle-satellite-label-disagreement}, distributions with label changes (N=15169) and no label changes (N=1106) were compared by computing the difference between view properties. From a t-test, four of the six view properties were significant: date of capture (p=9.528x10\textsuperscript{-8}), sun elevation (p=1.061x10\textsuperscript{-5}), sun azimuth (p=0.00017), and view azimuth (p=0.00039). The remaining two view properties were not significant, off nadir (p=0.9341) and incident angle (p=0.7525). 

\section{Discussion}
\label{sec:discussion}

The analysis, presented in Section \ref{sec:analysis}, finds that drone- and satellite-derived damage labels represent significantly different distributions and exhibit label disagreement, implicating concerns for CV/ML efforts in automated building damage assessments and their operational use. This section presents the limitations of the analysis in Section \ref{subsec:limitations}, discusses the properties of this analysis which separate it from the previous literature in Section \ref{subsec:analysis_properties}, describes the implications of the results of this analysis for the CV/ML community focused on building damage assessment in Section \ref{subsec:implications_cvml}, and highlights the risks, harms, and operational implications in Section \ref{subsec:risks_harms_ops}.

\subsection{Limitations}
\label{subsec:limitations}
This work has four limitations due to the limited availability of labeled imagery. First, this analysis does not consider the variety of potential resolutions, or scales, that exist between the satellite (64-30cm/px) and drone (1.94-4.67cm/px) scales. Second, this work focuses entirely on hurricane events as opposed to the variety of disasters that occur in practice (earthquakes, fires, tornadoes, etc.). Third, this work does not consider differing labels across drone views. Instead, this work is limited only to single drone views and multiple satellite views. Given the differences between the satellite and drone views, and the varying satellite views, it is likely that there is additional label variation in the drone views that is not being measured or considered. Finally, this work does not compare the drone- or satellite-derived labels with any ground-level labels or assessment of the buildings considered in this analysis.
While this type of comparison would be valuable, it is currently unclear how to fairly compare imagery-derived labels to ground-level assessments as the damage schemas used for aerial imagery \cite{gupta2019creating} and those used for ground-level damage assessment \cite{vickery2006hazus2,
vickery2006hazus1, grunthal1998european, emergency2002federal, kelman2003physical} have an undefined relationship. Nonetheless, the label disagreement observed between drone- and satellite-derived labels suggests that at least one of these label distributions will not align with ground-level labels at a distributional level, though it may align for individual labels.

\subsection{Analysis Properties}
\label{subsec:analysis_properties}
This analysis overcomes the limitations of prior literature by considering more than one order of magnitude more buildings, labeled using the same damage label schema and same building locations across three hurricanes. The most relevant prior work \cite{robinson2023rapid} considered label agreement between CV/ML model-sourced imagery-derived labels and human-sourced ground-level assessments of 830 buildings, but this was constrained by differing damage schemas between the imagery-derived labels, being a damage score between 0 and 100, and ground-level damage assessments, being the "FEMA building damage assessment scale" \cite{robinson2023rapid}.

This work analyzes 15,814 buildings observed in both drone and satellite imagery across hurricanes Harvey, Michael, and Ian. This represents 19.05 times more buildings than the previous literature (830 buildings \cite{robinson2023rapid}). It is expected that increasing the number of buildings considered in this analysis will lead to a better understanding of the label disagreement phenomenon.

The buildings analyzed in this audit were compared using the Joint Damage Scale \cite{gupta2019creating}. This meant that both drone- and satellite-derived labels could be compared directly as it was the label schema used in the CRASAR-U-DROIDs dataset \cite{manzini2024crasar} and the only building damage schema tailored for remote sensing.

Finally, this analysis provides labels for the same building polygon locations in both drone and satellite imagery, utilizing the same building polygons as the CRASAR-U-DROIDs dataset \cite{manzini2024crasar}. This was an intentional choice to avoid the localization uncertainty identified in \cite{robinson2023rapid}.

\subsection{Implications for Automated Building Damage Assessment}
\label{subsec:implications_cvml}
The fact that drone- and satellite-derived building damage assessment labels are significantly different distributions has three implications for how the CV/ML community builds damage assessment models. These implications all focus on the high degree of label disagreement observed in this analysis can induce noise in both the training and evaluation of CV/ML systems and in the communication of those model outputs in operational settings.

The first and largest implication is that the same label cannot be used across multiple views of the same building unless the developers are willing to tolerate label noise equivalent to the label disagreement rates that have been presented in this work. While the labels may be equivalent for certain buildings, this work provides evidence that this will not be true in all cases at a rate greater than statistical noise.
For example, it appears that should drone-derived labels be used to train a satellite damage assessment model, then such a model would experience a significant bias towards over-estimating damage compared to the damage that could be supported by evidence present in the imagery. The CV/ML community is starting to see remote-sensing datasets that depict the same scene from multiple sensors using the same label \cite{fang2024globally, garioud2024flair, d2023ai4boundaries}, and this work indicates that this should be avoided for building damage assessment unless the label noise described in this work is tolerable for a specific use case.

The second implication is that the damage label schemas currently used to label aerial imagery cannot handle label uncertainty observed in this work. Assigning a single label may not be appropriate, and depending on the imagery, it may be more appropriate to label a building's presence instead of a specific damage class.%For example, it may be unreasonable to distinguish the difference between major and minor damage at a spatial resolution greater than 1m/px; however, it may be possible to determine if the building is not destroyed. %This type of nuance should be captured in future iterations of building damage schemas.

The third and final implication is that continuing to optimize model performance on imagery-derived labels may not be appropriate if those imagery-derived labels misrepresent the distribution of ground-level conditions within the scene. 
It follows that further performance increases from imagery-derived labels may not add value to disaster response operations unless those performance increases are outside the label disagreement rates observed in this work. 
For example, although a 10\% gain in label agreement by a CV/ML model would normally be cause for celebration, the value of such an improvement is now questionable as it is less than the  29.02\% disagreement observed in this work.
Instead, it is argued that the CV/ML community should shift its efforts away from optimizing imagery-derived labels for building damage assessment and instead learn calibrated models that can correctly bound the ground-level damage labels.
%While ground-level labels are far harder to obtain and align with compared to the imagery-derived labels for building polygons used throughout the CV/ML for building damage assessment literature, 

\subsection{Risks, Harms \& Operational Implications}
\label{subsec:risks_harms_ops}
There are three potential risks and ethical implications that can lead to societal harm that arise from the findings in this analysis. These risks all stem from the fact that damage assessment systems are being trained and tested on a label distribution that may disagree with the distribution of labels they are being deployed against. These risks will continue to exist unless they are specifically addressed. 

%Despite training data being accurate based on the information the imagery...
The most substantive risk stems from the misclassification of damage, leading to a misallocation of resources or mismanagement of operations to the detriment of the public. Inaccurate classifications of damage from aerial imagery have led to harm before \cite{NASGCC:24}, and as CV/ML systems become more integrated into operations, they risk adversely affecting operations and societal outcomes unless systems are both performant and aligned with operational needs. The specific harms that can result from these errors are delayed response and/or aid to affected infrastructure or individuals and delayed and/or incorrect reimbursements to affected individuals for damage to life and property. Delays and errors during the response phase, such as these, create further downstream delays impeding recovery efforts \cite{haas1977reconstruction}.

%Due to noisy training data, predictions have higher levels of inaccuracy. An inaccurate prediction can lead to delayed response when being used by the emergency managers, inconsistent reimbursements derived from prediction results for the affected and risk to life and property due to insufficient response from the emergency managers <needs citations>

Separately from misclassification, there is the risk of acute rejection of model outputs as untrustworthy during ongoing operations. If this occurs in a real-world setting where reliance on model outputs is expected, this could result in an unmanageable data deluge, where humans would have to be rapidly retasked to inspect imagery to assess damage manually. Such a situation would delay decision-making, stress the human decision-makers and those tasked with inspecting imagery, and sow distrust in the underlying technology.

Finally, there is the risk that decision-makers would chronically reject and label these automated damage assessment technologies untrustworthy or unreliable after enough failures and errors. Such an outcome would represent a substantial obstacle to, or entirely block, the adoption of automated damage assessment systems in disaster response applications.

\section{Conclusion}
\label{sec:conclusion}
%Priya: Moved summary to topic paragraph
This work presents the first analysis of damage labels derived from drone and satellite imagery of coincident buildings and finds that damage label distributions significantly represent different distributions with label disagreement greater than 29.02\%, posing three potential risks and harms when deploying systems trained/evaluated on such data. 
Without appropriate management of the risk from these differing distributions, CV/ML practitioners may deliver incorrect damage assessments into disaster operations, which may lead to societal harm and reduced trust. The community must address this label disagreement and its implications for automated damage assessment systems to be trusted in practice. In response, this section provides four recommendations (Section \ref{subsec:conc_recommendations}) and finishes with a description of future work to be conducted (Section \ref{subsec:future_work})

%begins with a summary of the work presented in this document in Section \ref{subsec:conc_summary},

\subsection{Recommendations}
\label{subsec:conc_recommendations}
Based on the analysis in this work, four recommendations are presented to improve reliability and provide transparency to decision-makers when deploying CV/ML systems into practice. 

Recommendation \#1: \textit{Evaluate CV/ML systems with the appropriate ``ground truth” labels for the respective application}. All publicly available datasets for developing automated building damage assessments utilize imagery-derived labels. These labels likely represent a distribution significantly different from the distribution of ground-level labels. Depending on the application, it may or may not be appropriate to use imagery-derived labels to measure the performance. Evaluation of these damage assessment systems must utilize testing data that aligns as closely as possible with the intended use of the system and explicitly communicate the intended use case to non-technical decision-makers. 

Recommendation \#2: \textit{Frame automated damage assessment systems as a joint cognitive system instead of only a CV/ML problem}. Today, automated damage assessment systems are trained and evaluated based on their capability to detect damage. However, to realize any benefit from such a model, model outputs must be communicated to decision-makers who can leverage the information for operational benefit. So it is recommended that developers frame damage assessment systems as a joint cognitive system \cite{hollnagel2005joint, woods2006joint} between human decision-makers and automated damage assessment systems. Specifically focusing on how model outputs are communicated to decision-makers, how decision-makers use those model outputs to make decisions, and ultimately how those decisions impact outcomes.

Recommendation \#3: \textit{Damage assessment models must be calibrated to the label disagreements observed across views}. With the knowledge that imagery-derived building damage labels can vary across views, building damage assessment models must return outputs that are appropriately calibrated to this type of label variation and resulting uncertainty. Based on the amount and type of label disagreement observed in this work, it is argued that it is inappropriate to return a single label for a building based on a single view. Instead, damage assessment systems should return calibrated labels based on the range of views that may be available at a given time.

Recommendation \#4: \textit{Inform decision makers of the potential differences between aerial imagery-derived labels and ground-level assessments.}. Given the label disagreement observed in this work and the potential for real-world societal harm due to errors, presenting model outputs to decision-makers in operational contexts should be made with great care, and decision-makers must be informed of the potential for differences between aerial imagery-derived labels and ground-level assessments of building damage. This work shows that even if CV/ML models can achieve 0\% error on imagery-derived test set labels, that may represent as much as 29.02\% label error when measured against different sources. To this end, damage assessment models must be coupled with human oversight, with the knowledge that errors such as those described in this work may be present.

\subsection{Future Work}
\label{subsec:future_work}
Future efforts within this line of work will focus on further measuring disagreement and uncertainty in imagery-derived labels to effectively apply them in real-world operational contexts. To this end, there are three directions for future work. First, the drone- and satellite-derived labels will be compared with labels derived from crewed aircraft imagery. Second, a comparison with labels derived from ground-level assessments of buildings will be made to determine the label error of imagery-derived labels instead of label disagreement. Third, and finally, a mapping from aerial imagery-based damage scales to the damage scales commonly used for ground-level damage assessment using imagery-derived labels (drone, crewed aircraft, and satellite) will be developed.

\section*{Data Availability Statement}
\label{sec:data_availability}
All data associated with this effort can be found on Hugging Face under \url{https://huggingface.co/datasets/CRASAR/CRASAR-U-DROIDs} at commit ID 58f0d5ea2544dec8c126ac066e236943f26d0b7e. All code associated with this effort can be found on GitHub under repository \url{https://github.com/TManzini/NowYouSeeItNowYouDont}. 

\begin{acks}
The content of this manuscript is based upon work supported by the National
Science Foundation under AI Institute for Societal Decision
Making (AI-SDM), Award No. 2229881 and under “Datasets
for Uncrewed Aerial System (UAS) and Remote Responder
Performance from Hurricane Ian,” Award No. 2306453. The authors thank MAXAR for their willingness to release disaster imagery through their open data portal. The authors also thank the 83 annotators and their instructors from Ball High School and Rudder High School for their annotation efforts. Further thanks are given to Stephen Johnson, Raisa Karnik, Hasnat Md. Abdullah, Jeffery Hykin, James Lee, and Junsuk Kim for their helpful feedback on the annotation task, and to Ritwik Gupta for his guidance on the JDS and his feedback on the initial framing of this work.

\end{acks}

\bibliographystyle{ACM-Reference-Format}
\bibliography{main}


\begin{thebibliography}{41}

%%% the Web address.

\ifx \showCODEN    \undefined \def \showCODEN     #1{\unskip}     \fi
\ifx \showISBNx    \undefined \def \showISBNx     #1{\unskip}     \fi
\ifx \showISBNxiii \undefined \def \showISBNxiii  #1{\unskip}     \fi
\ifx \showISSN     \undefined \def \showISSN      #1{\unskip}     \fi
\ifx \showLCCN     \undefined \def \showLCCN      #1{\unskip}     \fi
\ifx \shownote     \undefined \def \shownote      #1{#1}          \fi
\ifx \showarticletitle \undefined \def \showarticletitle #1{#1}   \fi
\ifx \showURL      \undefined \def \showURL       {\relax}        \fi
% invisible to TeX
\providecommand\bibfield[2]{#2}
\providecommand\bibinfo[2]{#2}
\providecommand\natexlab[1]{#1}
\providecommand\showeprint[2][]{arXiv:#2}

\bibitem[max({[n.\,d.]})]%
        {maxar15}
 \bibinfo{year}{[n.\,d.]}\natexlab{}.
\newblock \bibinfo{booktitle}{\emph{Maxar Open Data}}.
\newblock
\urldef\tempurl%
\url{https://registry.opendata.aws/maxar-open-data/}
\showURL{%
\tempurl}


\bibitem[Mic(2021)]%
        {MicrosoftBuildingFootprints}
 \bibinfo{year}{2021}\natexlab{}.
\newblock \bibinfo{title}{Microsoft US Building Footprints}.
\newblock \bibinfo{howpublished}{\url{https://github.com/Microsoft/USBuildingFootprints}}.
\newblock


\bibitem[Lab(2024)]%
        {Labelbox2024}
 \bibinfo{year}{2024}\natexlab{}.
\newblock \bibinfo{booktitle}{\emph{Labelbox}}.
\newblock
\urldef\tempurl%
\url{https://labelbox.com}
\showURL{%
\tempurl}


\bibitem[Afonso et~al\mbox{.}(2019)]%
        {GFDRR:19}
\bibfield{author}{\bibinfo{person}{Stefanie Afonso}, \bibinfo{person}{Samir Belabbes}, \bibinfo{person}{Hélène~de Boissezon}, \bibinfo{person}{Jens Danzeglocke}, \bibinfo{person}{Andrew Eddy}, \bibinfo{person}{Joe Leitmann}, \bibinfo{person}{Mare Lo}, \bibinfo{person}{Rita Missal}, \bibinfo{person}{Françoise Villette}, \bibinfo{person}{Ricardo~Zapata Marti}, {and} \bibinfo{person}{Simona Zoffoli}.} \bibinfo{year}{2019}\natexlab{}.
\newblock \bibinfo{booktitle}{\emph{Use of EO Satellites in Support of Recovery from Major Disasters: Taking Stock and Moving Forward}}.
\newblock \bibinfo{type}{Report}. \bibinfo{institution}{Global Facility for Disaster Reduction and Recovery}.
\newblock


\bibitem[Agency(2021)]%
        {emergency2002federal}
\bibfield{author}{\bibinfo{person}{Federal Emergency~Management Agency}.} \bibinfo{year}{2021}\natexlab{}.
\newblock \bibinfo{title}{2021 Preliminary Damage Assessment Guide}.
\newblock


\bibitem[Akter and Wamba(2019)]%
        {AnkerAOR:19}
\bibfield{author}{\bibinfo{person}{Shahriar Akter} {and} \bibinfo{person}{Samuel~Fosso Wamba}.} \bibinfo{year}{2019}\natexlab{}.
\newblock \showarticletitle{Big data and disaster management: a systematic review and agenda for future research}.
\newblock \bibinfo{journal}{\emph{Annals of Operations Research}} \bibinfo{volume}{283}, \bibinfo{number}{1} (\bibinfo{year}{2019}), \bibinfo{pages}{939--959}.
\newblock
\showISSN{1572-9338}
\href{https://doi.org/10.1007/s10479-017-2584-2}{doi:\nolinkurl{10.1007/s10479-017-2584-2}}


\bibitem[Botzen et~al\mbox{.}(2019)]%
        {botzen:19}
\bibfield{author}{\bibinfo{person}{W.~J.~Wouter Botzen}, \bibinfo{person}{Olivier Deschenes}, {and} \bibinfo{person}{Mark Sanders}.} \bibinfo{year}{2019}\natexlab{}.
\newblock \showarticletitle{The Economic Impacts of Natural Disasters: A Review of Models and Empirical Studies}.
\newblock \bibinfo{journal}{\emph{Review of Environmental Economics and Policy}} \bibinfo{volume}{13}, \bibinfo{number}{2} (\bibinfo{year}{2019}), \bibinfo{pages}{167--188}.
\newblock
\href{https://doi.org/10.1093/reep/rez004}{doi:\nolinkurl{10.1093/reep/rez004}}


\bibitem[Chen et~al\mbox{.}(2025)]%
        {chen2025bright}
\bibfield{author}{\bibinfo{person}{Hongruixuan Chen}, \bibinfo{person}{Jian Song}, \bibinfo{person}{Olivier Dietrich}, \bibinfo{person}{Clifford Broni-Bediako}, \bibinfo{person}{Weihao Xuan}, \bibinfo{person}{Junjue Wang}, \bibinfo{person}{Xinlei Shao}, \bibinfo{person}{Yimin Wei}, \bibinfo{person}{Junshi Xia}, \bibinfo{person}{Cuiling Lan}, {et~al\mbox{.}}} \bibinfo{year}{2025}\natexlab{}.
\newblock \showarticletitle{BRIGHT: A globally distributed multimodal building damage assessment dataset with very-high-resolution for all-weather disaster response}.
\newblock \bibinfo{journal}{\emph{arXiv preprint arXiv:2501.06019}} (\bibinfo{year}{2025}).
\newblock


\bibitem[Chen et~al\mbox{.}(2018)]%
        {chen2018benchmark}
\bibfield{author}{\bibinfo{person}{Sean~Andrew Chen}, \bibinfo{person}{Andrew Escay}, \bibinfo{person}{Christopher Haberland}, \bibinfo{person}{Tessa Schneider}, \bibinfo{person}{Valentina Staneva}, {and} \bibinfo{person}{Youngjun Choe}.} \bibinfo{year}{2018}\natexlab{}.
\newblock \showarticletitle{Benchmark dataset for automatic damaged building detection from post-hurricane remotely sensed imagery}.
\newblock \bibinfo{journal}{\emph{arXiv preprint arXiv:1812.05581}} (\bibinfo{year}{2018}).
\newblock


\bibitem[Cheng et~al\mbox{.}(2021)]%
        {cheng2021dorianet}
\bibfield{author}{\bibinfo{person}{Chih-Shen Cheng}, \bibinfo{person}{Amir~H Behzadan}, {and} \bibinfo{person}{Arash Noshadravan}.} \bibinfo{year}{2021}\natexlab{}.
\newblock \showarticletitle{DoriaNET: A visual dataset from Hurricane Dorian for post-disaster building damage assessment}.
\newblock \bibinfo{journal}{\emph{DesignSafe-CI}} (\bibinfo{year}{2021}).
\newblock


\bibitem[d'Andrimont et~al\mbox{.}(2023)]%
        {d2023ai4boundaries}
\bibfield{author}{\bibinfo{person}{Rapha{\"e}l d'Andrimont}, \bibinfo{person}{Martin Claverie}, \bibinfo{person}{Pieter Kempeneers}, \bibinfo{person}{Davide Muraro}, \bibinfo{person}{Momchil Yordanov}, \bibinfo{person}{Devis Peressutti}, \bibinfo{person}{Matej Bati{\v{c}}}, {and} \bibinfo{person}{Fran{\c{c}}ois Waldner}.} \bibinfo{year}{2023}\natexlab{}.
\newblock \showarticletitle{AI4Boundaries: an open AI-ready dataset to map field boundaries with Sentinel-2 and aerial photography}.
\newblock \bibinfo{journal}{\emph{Earth System Science Data}} \bibinfo{volume}{15}, \bibinfo{number}{1} (\bibinfo{year}{2023}), \bibinfo{pages}{317--329}.
\newblock


\bibitem[Fang et~al\mbox{.}(2024)]%
        {fang2024globally}
\bibfield{author}{\bibinfo{person}{Chengyong Fang}, \bibinfo{person}{Xuanmei Fan}, \bibinfo{person}{Xin Wang}, \bibinfo{person}{Lorenzo Nava}, \bibinfo{person}{Hao Zhong}, \bibinfo{person}{Xiujun Dong}, \bibinfo{person}{Jixiao Qi}, {and} \bibinfo{person}{Filippo Catani}.} \bibinfo{year}{2024}\natexlab{}.
\newblock \showarticletitle{A globally distributed dataset of coseismic landslide mapping via multi-source high-resolution remote sensing images}.
\newblock \bibinfo{journal}{\emph{Earth System Science Data}} \bibinfo{volume}{16}, \bibinfo{number}{10} (\bibinfo{year}{2024}), \bibinfo{pages}{4817--4842}.
\newblock


\bibitem[Fernandes et~al\mbox{.}(2018)]%
        {fernandes2018quantitative}
\bibfield{author}{\bibinfo{person}{Odair Fernandes}, \bibinfo{person}{Robin Murphy}, \bibinfo{person}{Justin Adams}, {and} \bibinfo{person}{David Merrick}.} \bibinfo{year}{2018}\natexlab{}.
\newblock \showarticletitle{Quantitative data analysis: CRASAR small unmanned aerial systems at hurricane Harvey}. In \bibinfo{booktitle}{\emph{2018 IEEE International Symposium on Safety, Security, and Rescue Robotics (SSRR)}}. IEEE, \bibinfo{pages}{1--6}.
\newblock


\bibitem[Fernandes et~al\mbox{.}(2019)]%
        {fernandes2019quantitative}
\bibfield{author}{\bibinfo{person}{Odair Fernandes}, \bibinfo{person}{Robin Murphy}, \bibinfo{person}{David Merrick}, \bibinfo{person}{Justin Adams}, \bibinfo{person}{Laura Hart}, {and} \bibinfo{person}{Jarrett Broder}.} \bibinfo{year}{2019}\natexlab{}.
\newblock \showarticletitle{Quantitative data analysis: Small unmanned aerial systems at hurricane michael}. In \bibinfo{booktitle}{\emph{2019 IEEE international symposium on safety, security, and rescue robotics (SSRR)}}. IEEE, \bibinfo{pages}{116--117}.
\newblock


\bibitem[Fujita et~al\mbox{.}(2017)]%
        {fujita2017damage}
\bibfield{author}{\bibinfo{person}{Aito Fujita}, \bibinfo{person}{Ken Sakurada}, \bibinfo{person}{Tomoyuki Imaizumi}, \bibinfo{person}{Riho Ito}, \bibinfo{person}{Shuhei Hikosaka}, {and} \bibinfo{person}{Ryosuke Nakamura}.} \bibinfo{year}{2017}\natexlab{}.
\newblock \showarticletitle{Damage detection from aerial images via convolutional neural networks}. In \bibinfo{booktitle}{\emph{2017 Fifteenth IAPR international conference on machine vision applications (MVA)}}. IEEE, \bibinfo{pages}{5--8}.
\newblock


\bibitem[Garc{\'\i}a~Franceschini(2021)]%
        {garcia2021computer}
\bibfield{author}{\bibinfo{person}{Ren{\'e}~Andr{\'e}s Garc{\'\i}a~Franceschini}.} \bibinfo{year}{2021}\natexlab{}.
\newblock \emph{\bibinfo{title}{Computer vision-based post-disaster needs assessment from low altitude aerial imagery}}.
\newblock \bibinfo{thesistype}{Ph.\,D. Dissertation}. \bibinfo{school}{Massachusetts Institute of Technology}.
\newblock


\bibitem[Garioud et~al\mbox{.}(2024)]%
        {garioud2024flair}
\bibfield{author}{\bibinfo{person}{Anatol Garioud}, \bibinfo{person}{Nicolas Gonthier}, \bibinfo{person}{Loic Landrieu}, \bibinfo{person}{Apolline De~Wit}, \bibinfo{person}{Marion Valette}, \bibinfo{person}{Marc Poup{\'e}e}, \bibinfo{person}{S{\'e}bastien Giordano}, {et~al\mbox{.}}} \bibinfo{year}{2024}\natexlab{}.
\newblock \showarticletitle{FLAIR: a country-scale land cover semantic segmentation dataset from multi-source optical imagery}.
\newblock \bibinfo{journal}{\emph{Advances in Neural Information Processing Systems}}  \bibinfo{volume}{36} (\bibinfo{year}{2024}).
\newblock


\bibitem[Gr{\"u}nthal(1998)]%
        {grunthal1998european}
\bibfield{author}{\bibinfo{person}{Gottfried Gr{\"u}nthal}.} \bibinfo{year}{1998}\natexlab{}.
\newblock \showarticletitle{European macroseismic scale 1998 (EMS-98)}.
\newblock  (\bibinfo{year}{1998}).
\newblock


\bibitem[Gupta et~al\mbox{.}(2019)]%
        {gupta2019creating}
\bibfield{author}{\bibinfo{person}{Ritwik Gupta}, \bibinfo{person}{Bryce Goodman}, \bibinfo{person}{Nirav Patel}, \bibinfo{person}{Ricky Hosfelt}, \bibinfo{person}{Sandra Sajeev}, \bibinfo{person}{Eric Heim}, \bibinfo{person}{Jigar Doshi}, \bibinfo{person}{Keane Lucas}, \bibinfo{person}{Howie Choset}, {and} \bibinfo{person}{Matthew Gaston}.} \bibinfo{year}{2019}\natexlab{}.
\newblock \showarticletitle{Creating xBD: A dataset for assessing building damage from satellite imagery}. In \bibinfo{booktitle}{\emph{Proceedings of the IEEE/CVF conference on computer vision and pattern recognition workshops}}. \bibinfo{pages}{10--17}.
\newblock


\bibitem[Gyegyiri et~al\mbox{.}(2024)]%
        {IanBD}
\bibfield{author}{\bibinfo{person}{Joseph Gyegyiri}, \bibinfo{person}{Hongbo Su}, {and} \bibinfo{person}{Ajay Thapa}.} \bibinfo{year}{2024}\natexlab{}.
\newblock \bibinfo{title}{Ian-BD: pre- and post-disaster high-resolution aerial imagery for building damage assessment from Hurricane Ian.}
\newblock
\href{https://doi.org/10.17603/DS2-2N0D-4S77}{doi:\nolinkurl{10.17603/DS2-2N0D-4S77}}


\bibitem[Haas et~al\mbox{.}(1977)]%
        {haas1977reconstruction}
\bibfield{author}{\bibinfo{person}{J~Eugene Haas}, \bibinfo{person}{Robert~W Kates}, {and} \bibinfo{person}{Martyn~J Bowden}.} \bibinfo{year}{1977}\natexlab{}.
\newblock \showarticletitle{Reconstruction following disaster}.
\newblock In \bibinfo{booktitle}{\emph{Reconstruction following disaster}}. \bibinfo{pages}{331--331}.
\newblock


\bibitem[Hollnagel and Woods(2005)]%
        {hollnagel2005joint}
\bibfield{author}{\bibinfo{person}{Erik Hollnagel} {and} \bibinfo{person}{David~D Woods}.} \bibinfo{year}{2005}\natexlab{}.
\newblock \bibinfo{booktitle}{\emph{Joint cognitive systems: Foundations of cognitive systems engineering}}.
\newblock \bibinfo{publisher}{CRC press}.
\newblock


\bibitem[Kelman(2003)]%
        {kelman2003physical}
\bibfield{author}{\bibinfo{person}{Ilan Kelman}.} \bibinfo{year}{2003}\natexlab{}.
\newblock \emph{\bibinfo{title}{Physical flood vulnerability of residential properties in coastal, eastern England}}.
\newblock \bibinfo{thesistype}{Ph.\,D. Dissertation}. \bibinfo{school}{University of Cambridge Cambridge, United Kingdom}.
\newblock


\bibitem[Lee et~al\mbox{.}(2022)]%
        {IdaBDDataset}
\bibfield{author}{\bibinfo{person}{Cheng-Chun Lee}, \bibinfo{person}{Navjot Kaur}, \bibinfo{person}{Ali Mahdavi-Amiri}, {and} \bibinfo{person}{Ali Mostafavi}.} \bibinfo{year}{2022}\natexlab{}.
\newblock \bibinfo{title}{Ida-BD: pre- and post-disaster high-resolution satellite imagery for building damage assessment from Hurricane Ida}.
\newblock
\href{https://doi.org/10.17603/DS2-PD9H-6K05}{doi:\nolinkurl{10.17603/DS2-PD9H-6K05}}


\bibitem[Lin et~al\mbox{.}(2023)]%
        {haitiBRDDataset}
\bibfield{author}{\bibinfo{person}{Szu-Yun Lin}, \bibinfo{person}{Marisa Edocia}, \bibinfo{person}{Fang~Jung Tsai}, \bibinfo{person}{Lien~an Chen}, {and} \bibinfo{person}{Wen-Ni Kuo}.} \bibinfo{year}{2023}\natexlab{}.
\newblock \bibinfo{title}{HaitiBRD: A labeled satellite imagery dataset for building and road damage assessment of the 2010 Haiti earthquake}.
\newblock
\href{https://doi.org/10.17603/DS2-FQAT-4V02}{doi:\nolinkurl{10.17603/DS2-FQAT-4V02}}


\bibitem[{Liu} et~al\mbox{.}(2019)]%
        {liuLargeScale2019}
\bibfield{author}{\bibinfo{person}{J. {Liu}}, \bibinfo{person}{D. {Strohschein}}, \bibinfo{person}{S. {Samsi}}, {and} \bibinfo{person}{A. {Weinert}}.} \bibinfo{year}{2019}\natexlab{}.
\newblock \showarticletitle{Large Scale Organization and Inference of an Imagery Dataset for Public Safety}. In \bibinfo{booktitle}{\emph{2019 IEEE High Performance Extreme Computing Conference (HPEC)}}. \bibinfo{pages}{1--6}.
\newblock
\showISSN{2377-6943}
\href{https://doi.org/10.1109/HPEC.2019.8916437}{doi:\nolinkurl{10.1109/HPEC.2019.8916437}}


\bibitem[Lozano and Tien(2023)]%
        {LozanoIJDRR:23}
\bibfield{author}{\bibinfo{person}{Jorge-Mario Lozano} {and} \bibinfo{person}{Iris Tien}.} \bibinfo{year}{2023}\natexlab{}.
\newblock \showarticletitle{Data collection tools for post-disaster damage assessment of building and lifeline infrastructure systems}.
\newblock \bibinfo{journal}{\emph{International Journal of Disaster Risk Reduction}}  \bibinfo{volume}{94} (\bibinfo{year}{2023}), \bibinfo{pages}{103819}.
\newblock
\showISSN{2212-4209}
\href{https://doi.org/10.1016/j.ijdrr.2023.103819}{doi:\nolinkurl{10.1016/j.ijdrr.2023.103819}}


\bibitem[Manzini et~al\mbox{.}(2023a)]%
        {manzini2023quantitative}
\bibfield{author}{\bibinfo{person}{Thomas Manzini}, \bibinfo{person}{Robin Murphy}, {and} \bibinfo{person}{David Merrick}.} \bibinfo{year}{2023}\natexlab{a}.
\newblock \showarticletitle{Quantitative Data Analysis: CRASAR Small Unmanned Aerial Systems at Hurricane Ian}. In \bibinfo{booktitle}{\emph{2023 IEEE International Symposium on Safety, Security, and Rescue Robotics (SSRR)}}. IEEE, \bibinfo{pages}{7--12}.
\newblock


\bibitem[Manzini et~al\mbox{.}(2023c)]%
        {manzini2023wireless}
\bibfield{author}{\bibinfo{person}{Thomas Manzini}, \bibinfo{person}{Robin Murphy}, \bibinfo{person}{David Merrick}, {and} \bibinfo{person}{Justin Adams}.} \bibinfo{year}{2023}\natexlab{c}.
\newblock \showarticletitle{Wireless Network Demands of Data Products from Small Uncrewed Aerial Systems at Hurricane Ian}. In \bibinfo{booktitle}{\emph{2023 IEEE/RSJ International Conference on Intelligent Robots and Systems (IROS)}}. IEEE, \bibinfo{pages}{9941--9946}.
\newblock


\bibitem[Manzini et~al\mbox{.}(2023b)]%
        {manzini2023harnessing}
\bibfield{author}{\bibinfo{person}{Thomas Manzini}, \bibinfo{person}{Robin~R Murphy}, \bibinfo{person}{Eric Heim}, \bibinfo{person}{Caleb Robinson}, \bibinfo{person}{Guido Zarrella}, {and} \bibinfo{person}{Ritwik Gupta}.} \bibinfo{year}{2023}\natexlab{b}.
\newblock \showarticletitle{Harnessing AI and robotics in humanitarian assistance and disaster response}.
\newblock \bibinfo{journal}{\emph{Science robotics}} \bibinfo{volume}{8}, \bibinfo{number}{80} (\bibinfo{year}{2023}), \bibinfo{pages}{eadj2767}.
\newblock


\bibitem[Manzini et~al\mbox{.}(2024)]%
        {manzini2024crasar}
\bibfield{author}{\bibinfo{person}{Thomas Manzini}, \bibinfo{person}{Priyankari Perali}, \bibinfo{person}{Raisa Karnik}, {and} \bibinfo{person}{Robin Murphy}.} \bibinfo{year}{2024}\natexlab{}.
\newblock \showarticletitle{CRASAR-U-DROIDs: A Large Scale Benchmark Dataset for Building Alignment and Damage Assessment in Georectified sUAS Imagery}.
\newblock \bibinfo{journal}{\emph{arXiv preprint arXiv:2407.17673}} (\bibinfo{year}{2024}).
\newblock


\bibitem[National Academies~of Sciences and Medicine(2024)]%
        {NASGCC:24}
\bibfield{author}{\bibinfo{person}{Engineering National Academies~of Sciences} {and} \bibinfo{person}{Medicine}.} \bibinfo{year}{2024}\natexlab{}.
\newblock \bibinfo{booktitle}{\emph{Compounding Disasters in Gulf Coast Communities 2020-2021: Impacts, Findings, and Lessons Learned}}.
\newblock \bibinfo{publisher}{The National Academies Press}, \bibinfo{address}{Washington, DC}. 296 pages.
\newblock
\showISBNx{978-0-309-70716-9}
\href{https://doi.org/doi:10.17226/27170}{doi:\nolinkurl{doi:10.17226/27170}}


\bibitem[Pi et~al\mbox{.}(2020)]%
        {Pi2020}
\bibfield{author}{\bibinfo{person}{Yalong Pi}, \bibinfo{person}{Nipun~D. Nath}, {and} \bibinfo{person}{Amir~H. Behzadan}.} \bibinfo{year}{2020}\natexlab{}.
\newblock \showarticletitle{Convolutional neural networks for object detection in aerial imagery for disaster response and recovery}.
\newblock \bibinfo{journal}{\emph{Advanced Engineering Informatics}}  \bibinfo{volume}{43} (\bibinfo{year}{2020}), \bibinfo{pages}{101009}.
\newblock


\bibitem[Rahnemoonfar et~al\mbox{.}(2022)]%
        {rahnemoonfar2022rescuenet}
\bibfield{author}{\bibinfo{person}{Maryam Rahnemoonfar}, \bibinfo{person}{Tashnim Chowdhury}, {and} \bibinfo{person}{Robin Murphy}.} \bibinfo{year}{2022}\natexlab{}.
\newblock \showarticletitle{RescueNet: A high resolution UAV semantic segmentation benchmark dataset for natural disaster damage assessment}.
\newblock \bibinfo{journal}{\emph{arXiv preprint arXiv:2202.12361}} (\bibinfo{year}{2022}).
\newblock


\bibitem[Rahnemoonfar et~al\mbox{.}(2021)]%
        {rahnemoonfar2021floodnet}
\bibfield{author}{\bibinfo{person}{Maryam Rahnemoonfar}, \bibinfo{person}{Tashnim Chowdhury}, \bibinfo{person}{Argho Sarkar}, \bibinfo{person}{Debvrat Varshney}, \bibinfo{person}{Masoud Yari}, {and} \bibinfo{person}{Robin~Roberson Murphy}.} \bibinfo{year}{2021}\natexlab{}.
\newblock \showarticletitle{Floodnet: A high resolution aerial imagery dataset for post flood scene understanding}.
\newblock \bibinfo{journal}{\emph{IEEE Access}}  \bibinfo{volume}{9} (\bibinfo{year}{2021}), \bibinfo{pages}{89644--89654}.
\newblock


\bibitem[Robinson et~al\mbox{.}(2023)]%
        {robinson2023rapid}
\bibfield{author}{\bibinfo{person}{Caleb Robinson}, \bibinfo{person}{Simone~Fobi Nsutezo}, \bibinfo{person}{Anthony Ortiz}, \bibinfo{person}{Tina Sederholm}, \bibinfo{person}{Rahul Dodhia}, \bibinfo{person}{Cameron Birge}, \bibinfo{person}{Kasie Richards}, \bibinfo{person}{Kris Pitcher}, \bibinfo{person}{Paulo Duarte}, {and} \bibinfo{person}{Juan M~Lavista Ferres}.} \bibinfo{year}{2023}\natexlab{}.
\newblock \showarticletitle{Rapid building damage assessment workflow: An implementation for the 2023 Rolling Fork, Mississippi tornado event}. In \bibinfo{booktitle}{\emph{Proceedings of the IEEE/CVF International Conference on Computer Vision}}. \bibinfo{pages}{3760--3764}.
\newblock


\bibitem[Sadek et~al\mbox{.}(2021)]%
        {sadek2021engineering}
\bibfield{author}{\bibinfo{person}{S Sadek}, \bibinfo{person}{M Dabaghi}, \bibinfo{person}{I Elhajj}, \bibinfo{person}{P Zimmaro}, \bibinfo{person}{YMA Hashash}, \bibinfo{person}{SH Yun}, \bibinfo{person}{TM O'Donnell}, \bibinfo{person}{JP Stewart}, {et~al\mbox{.}}} \bibinfo{year}{2021}\natexlab{}.
\newblock \showarticletitle{Engineering impacts of the August 4, 2020 Port of Beirut, Lebanon explosion}.
\newblock  (\bibinfo{year}{2021}).
\newblock


\bibitem[Vickery et~al\mbox{.}(2006a)]%
        {vickery2006hazus1}
\bibfield{author}{\bibinfo{person}{Peter~J Vickery}, \bibinfo{person}{Jason Lin}, \bibinfo{person}{Peter~F Skerlj}, \bibinfo{person}{Lawrence~A Twisdale~Jr}, {and} \bibinfo{person}{Kevin Huang}.} \bibinfo{year}{2006}\natexlab{a}.
\newblock \showarticletitle{HAZUS-MH hurricane model methodology. I: Hurricane hazard, terrain, and wind load modeling}.
\newblock \bibinfo{journal}{\emph{Natural Hazards Review}} \bibinfo{volume}{7}, \bibinfo{number}{2} (\bibinfo{year}{2006}), \bibinfo{pages}{82--93}.
\newblock


\bibitem[Vickery et~al\mbox{.}(2006b)]%
        {vickery2006hazus2}
\bibfield{author}{\bibinfo{person}{Peter~J Vickery}, \bibinfo{person}{Peter~F Skerlj}, \bibinfo{person}{Jason Lin}, \bibinfo{person}{Lawrence~A Twisdale~Jr}, \bibinfo{person}{Michael~A Young}, {and} \bibinfo{person}{Francis~M Lavelle}.} \bibinfo{year}{2006}\natexlab{b}.
\newblock \showarticletitle{HAZUS-MH hurricane model methodology. II: Damage and loss estimation}.
\newblock \bibinfo{journal}{\emph{Natural Hazards Review}} \bibinfo{volume}{7}, \bibinfo{number}{2} (\bibinfo{year}{2006}), \bibinfo{pages}{94--103}.
\newblock


\bibitem[Woods and Hollnagel(2006)]%
        {woods2006joint}
\bibfield{author}{\bibinfo{person}{David~D Woods} {and} \bibinfo{person}{Erik Hollnagel}.} \bibinfo{year}{2006}\natexlab{}.
\newblock \bibinfo{booktitle}{\emph{Joint cognitive systems: Patterns in cognitive systems engineering}}.
\newblock \bibinfo{publisher}{CRC press}.
\newblock


\bibitem[Zhu et~al\mbox{.}(2021)]%
        {zhu2021msnet}
\bibfield{author}{\bibinfo{person}{Xiaoyu Zhu}, \bibinfo{person}{Junwei Liang}, {and} \bibinfo{person}{Alexander Hauptmann}.} \bibinfo{year}{2021}\natexlab{}.
\newblock \showarticletitle{Msnet: A multilevel instance segmentation network for natural disaster damage assessment in aerial videos}. In \bibinfo{booktitle}{\emph{Proceedings of the IEEE/CVF winter conference on applications of computer vision}}. \bibinfo{pages}{2023--2032}.
\newblock


\end{thebibliography}

%\appendix

\end{document}